\documentclass[transmag]{IEEEtran}

\usepackage[hyphens]{url}
\usepackage{cite}
\usepackage{amsmath,amssymb,amsfonts}
\usepackage{graphicx}
\usepackage{xcolor}
\usepackage[inline]{enumitem}
\usepackage{float}
\restylefloat{table}
\begin{document}

\title{State-of-the-art Techniques in Deep Edge Intelligence}


\author{\IEEEauthorblockN{Ahnaf Hannan Lodhi}
\IEEEauthorblockA{\textit{Department of Computer Engineering} \\
\textit{Ko\c{c} University}\\
Istanbul, Turkey \\
alodhi18@ku.edu.tr}
\and
\IEEEauthorblockN{Bar{\i}\c{s} Akg\"{u}n}
\IEEEauthorblockA{\textit{Department of Computer Engineering} \\
\textit{Ko\c{c} University}\\
Istanbul, Turkey \\
baakgun@ku.edu.tr}
\and
\IEEEauthorblockN{\"{O}znur \"{O}zkasap}
\IEEEauthorblockA{\textit{Department of Computer Engineering} \\
\textit{Ko\c{c} University}\\
Istanbul, Turkey \\
oozkasap@ku.edu.tr}
}

\maketitle
\footnote{“This work has been submitted to the IEEE for possible publication. Copyright may be transferred without notice, after which this version may no longer be accessible.”}
\begin{abstract}

The potential held by the gargantuan volumes of data being generated across networks worldwide has been truly unlocked by machine learning techniques and more recently Deep Learning. The advantages offered by the latter have seen it rapidly becoming a framework of choice for various applications. However, the centralization of computational resources and the need for data aggregation have long been limiting factors in the democratization of Deep Learning applications. Edge Computing is an emerging paradigm that aims to utilize the hitherto untapped processing resources available at the network periphery. Edge Intelligence (EI) has quickly emerged as a powerful alternative to enable learning using the concepts of Edge Computing. Deep Learning-based Edge Intelligence or Deep Edge Intelligence (DEI) lies in this rapidly evolving domain. In this article, we provide an overview of the major constraints in operationalizing DEI. The major research avenues in DEI have been consolidated under Federated Learning, Distributed Computation, Compression Schemes and Conditional Computation. We also present some of the prevalent challenges and highlight prospective research avenues.
\end{abstract}

\section{Introduction}

The last decade has witnessed a remarkable transcendence of data and information-centric services in all spheres of the global domain. The availability of powerful computational resources and the ascendancy of learning systems have further led to unprecedented growth in data-generating devices and sensors. It has been estimated that close to 29 billion devices would be connected to the Internet by 2023  \cite{cisco} with the data traffic expected to reach 131 Exabytes (EB) by the end of 2024 \cite{ericsson}. Furthermore, the requirements for 6G aiming for data rates of approximately 1Tbps per user \cite{Oulu} have further reinforced a growing realization that the traditional centralized/cloud systems would be unable to efficiently manage the accompanying computation requirements.\\ 

The ability to imbue systems with intelligence is at the forefront of this technological revolution. Conventional machine learning tasks have rapidly found applications in multiple domains. Simultaneously, Deep Learning (DL) has risen meteorically through the last decade with unparalleled performance primarily in Computer Vision and Natural Language Processing (NLP) fields. However, the performance offered by Deep Learning systems comes with significant computation and memory costs in addition to massive data requirements. As of writing this article, the current State-Of-The-Art (SOTA) for image classification in the ImageNet Large Scale Visual Recognition Challenge (ILSVRC) is the FixEfficientNet-L2 \cite{touvron2020fixing} which achieves a Top-5 accuracy of 98\% using 480M parameters. However, providing similar performance in related applications at the user-end is currently constrained by limited computational resources exacerbated by collecting, communicating and storing the required amount of data.\\ 

The paradigm shift in the nature of networked services and the transformation of the connected devices coupled with the distributed nature of data requires a decentralized approach for extracting maximum learning benefits. To avoid overwhelming the network and data servers, the computational load must be moved at or closer to the network edge. Edge Computing \cite{shi2016edge} offers a potentially powerful solution to this problem. The edge computing framework aims to leverage distributed computing concepts to alleviate the computational load from the network core benefiting from processing power available close to the network edge. The confluence of Artificial Intelligence (AI) and Edge Computing results in Edge Intelligence (EI). Application of Deep Learning to achieve Edge Intelligence offers the added benefit of employing raw data without the considerable feature engineering and pre-processing overhead. The computation power of the elements closer to the edge network offers a powerful alternative to centralized computing albeit in a distributed manner. Successful exploitation may result in elements of cloud services being shifted in close proximity to the data sources ensuring better data security as well as reduced load on the network backbone.\\

Our contribution with this article is an attempt at formalizing the key constraints which must be addressed for realizing efficient DEI applications. To the best of our knowledge, these constraints have been disparately discussed while 'Device Disparity' and 'Inference Transparency' have not been formally considered previously. We further unify broad research avenues and classify the work under these categories to provide a concise overview of the evolution of Deep Edge Intelligence. Lastly, we attempt to identify challenges and research directions not only related to the implementation of DEI but also present some potential learning schemes which might be highly suitable for this rapidly evolving domain. To summarize, the unique highlights of this article are as follows:
\begin{enumerate}[label = \alph*)]
    \item We consolidate the operational constraints for Deep Learning-based Edge Intelligence to reflect aspects not only related to Edge Intelligence but also Deep Learning.
    \item We classify the avenues of research progression in Deep Learning-based Edge Intelligence. Our aim with this categorization is to provide a broad basis for the current and future research themes.
    \item We identify key areas which offer the potential for prospective research uniquely suited for optimal application of DEI.\\  
\end{enumerate}

With this article, we aim to elaborate on the evolution of research directions for Edge Intelligence (EI) in the presence of the prevalent challenges. Starting from the elucidating on the concepts of EI and key drivers behind Deep Edge Intelligence (DEI) in Section-\ref{EI}, we present the latest perspectives for achieving DEI in Section-\ref{enabling}. Finally in Sections-\ref{open}, we provide some open challenges and prospective research directions to achieve robust and efficient learning paradigms for DEI. This is followed by Section-\ref{conclusion} in which we summarize the details of this article along with some thoughts to spur constructive discussion on the topic.   

\section{Edge Intelligence}\label{EI}
The true potential of Artificial Intelligence (AI) is unlocked in a connected/networked domain where intelligent services can be extended simultaneously to a large scale of users. Some of the more applications are based on but not limited to facial recognition for smart surveillance, object recognition, language translation, sentiment analysis, and load prediction. Sufficient to say that Machine Learning has also been deployed as the first line of detection of the novel Coronavirus–caused respiratory disease (COVID-19) \cite{ting2020digital} at major transit points based on thermal imaging. Deployment of isolated or centralized processing clusters offers an inefficient solution due to resource wastage, prohibitive costs with the major disadvantage being that of the distributed nature of data itself. The development of a framework that supports distributed operation and access to remote data is an essential requirement for the rapidly transforming applications. 
\subsection{Edge Computing}
Edge Computing is the emerging domain being developed to address the limitations associated with cloud computing. This paradigm aims at shifting the computation load towards the outer edges of the network utilizing the devices at the data origin and their geographical proximity. The motivation behind the development of this domain is two-fold: It aims to reduce the computational and communication load from the network core while enabling dynamic resource allocation for applications in cyber-physical systems such as industrial IoT, smart buildings and grids, autonomous transportation, and remote healthcare \cite{Chen_Zhang_Maharjan_Alam_Wu_2019}. 
    \begin{figure}[htbp]
    \centerline{\includegraphics[scale=0.13]{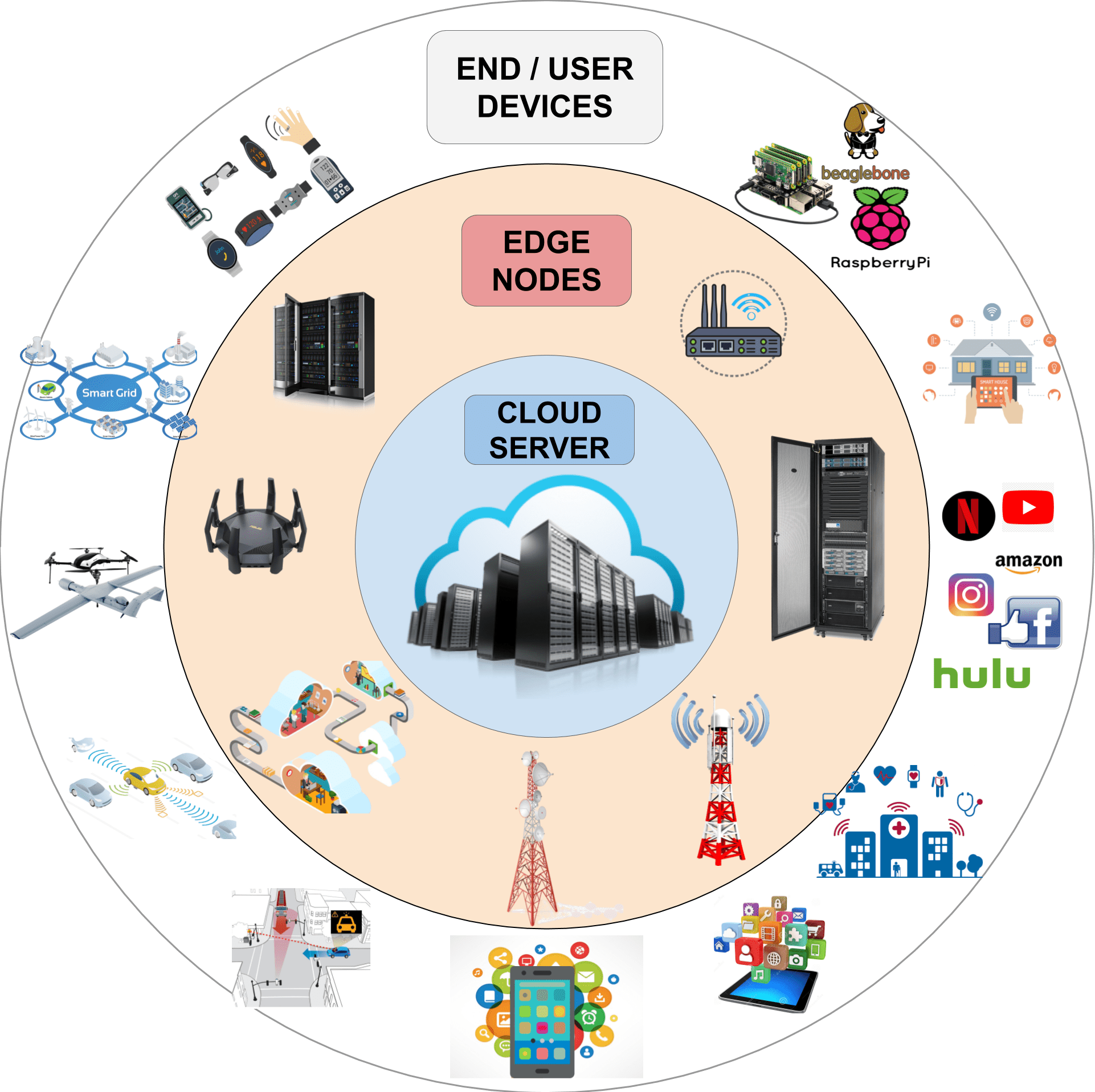}}
    \caption{Levels of Network Hierarchy: \emph{a) End / User device:} Devices/sensors at the origin of data, \emph{b) Edge Nodes:} Intermediate network devices connecting end devices to the network core, \emph{c) Cloud Servers:} Hub of computation and decision making in existing networks. The two outer levels are jointly referred to as \emph{"Edge Levels"} and are the focus of Edge Computing.}
    \label{fig:network}
    \end{figure}
Conventional networks can be characterized by reduced computational power in the elements at the fringes of the network. However, the addition of devices at the outer levels has also significantly outpaced the increase of processing power at the network core \cite{cisco}. Edge Computing has emerged to support this dramatic increase in resource requirements by leveraging the untapped potential away from the enterprise data centers. Processing power is obtained by a collaborative operation between various entities at the network edge including the user devices, mobile-based stations and gateways and access points.
\subsection{Network Hierarchy}
Due to the relatively early stages of development, Edge Computing offers a multitude of terminologies to depict various stages of the network. In general, however, a conventional network can be categorized into three hierarchical levels from the perspective of Edge Intelligence (EI) as depicted in Fig-\ref{fig:network}:
\begin{enumerate}
    \item \textbf{End or user devices} lie at the outermost levels of the network. These are the set of elements that are themselves responsible for data generation or house applications which do. These include but are not limited to mobile devices including smartphones, tablets and smart wearables, IoT devices, sensors for smart grids, homes and healthcare, and autonomous cars and drones. These devices are characterized by less computing power compared to other devices on the network. 
    \item \textbf{Edge server or nodes} comprise of mobile base stations, gateways, access points (APs), micro data-centers, etc. existing at the intermediate level between the Edge Devices and the Cloud/Data server level. These are established in proximity of the data sources having comparatively higher computational and storage capacity than user devices but still considerably lower capacity than the cloud level. \textit{Combined with the end devices, these two levels can be thought to jointly form the 'Edge Level'.}
    \item \textbf{Cloud or data server} level is the innermost level in the network hierarchy possessing maximal computation and storage ability. The processing is carried out at this level in conventional networks and settings. However, the remoteness of Cloud level from Edge devices incurs a considerable communication overhead, latency as well as potential privacy issues. 
   \end{enumerate}
\subsection {Edge Intelligence}
Edge Intelligence (EI) refers to the utilization of Artificial Intelligence (AI) paradigm in Edge Computing Scenarios. Traditionally intelligent inferences are generated at the cloud level, which required data aggregation before undertaking the learning process. Edge Intelligence, on the contrary, relies on establishing the AI support away from the cloud level imparting a certain degree of intelligence at the network edge. Edge Intelligence aims at maximally offloading the learning and inference computations to the edge level thus alleviating resource demands at the cloud level. In general Edge Intelligence requires a multi-disciplinary approach using knowledge and efficient practices from fields including but not limited to AI, Computer Architecture, Embedded systems, Compressive sensing and Distributed Systems, etc. for optimal performance. 
 
\subsection{Operational Constraints} \label{constraints}
Adopting distributed computing brings forth its own set of challenges for optimal operation including data sharing, security, latency, etc. Additionally, the typical learning scenarios have so far utilized centralized frameworks unifying both data and computing resources at one single entity in the form of servers or clusters. Edge Intelligence, on the contrary, aims to exploit the significantly untapped potential of the billions of elements at the network edge. The current research to achieve this goal is primarily driven by the following factors \cite{Chen_Ran_2019}, \cite{Wang_Han_Leung_Niyato_Yan_Chen_2020}:
\begin{enumerate}
    \item \textit{\textbf{Cost:}} Any form of decentralized computation results in major costs related to communication, energy, processing and memory. 
    \begin{itemize}
        \item \textit{Communication Costs:} Remote computations require data to be exchanged between various distributed elements introducing not only the cost of data communication but the associated overhead costs in already congested networks. The services at the end devices, whether provided by mobile applications over cellular networks or IoT networks are increasingly resorting to provide improved user experiences as well information. The demand for immersive Quality of Experience (QoE) using Augmented Reality / Virtual Reality (AR/VR) alone is expected to result in an 8-fold increase in data traffic. \cite{Taleb_Samdanis_Mada_Flinck_Dutta_Sabella_2017}. Furthermore, while the emerging networks are offering higher speeds, legacy networks would face increasingly difficult prospects for supporting such services. Incorporating Edge Intelligence in such an environment would thus be associated with its communication costs. 
        
         \item \textit{Energy:} A considerable majority of nodes at the edge level, whether the user devices or the edge nodes often operate with limited energy budget. As opposed to a cloud server where energy constraints are considered for economical operation, mobile devices at the user end cease operation if they exceed their energy constraints. Thus, all aspects of the learning process, from training to inference must respect these energy limitations.

        \item \textit{Processing and Memory:} Deep Learning models require considerable processing and memory resources to extract and learn the deep representations of the data. In addition to a dearth of processing power at the edge level, managing sufficient resources for deep learning models to run in the presence of numerous competing services is a major challenge. In general, deeper networks more often outperform shallow networks and thus higher performance requires larger storage requirements.  
       
        \end{itemize}

    \item \textit{\textbf{Latency:}} Time-critical applications require real-time or near-real-time inferences. For example, translation of conversations from one language to another, object segmentation in photos or analysis, fusion and logical inference of data carried out by the sensors of autonomous vehicles especially self-driving cars are some applications where the delay between the input and inference can result in seriously compromised performance. Additionally, offboarding data for remote computation imposed additional time costs due to communication to deeper levels of the network. \cite{satyanarayanan2017emergence} presented effects of proximity for a face detection task running on Amazon Web Services indicating a 50\% task completion rate between 200-600ms depending on the server location.

    \item \textit{\textbf{Scalability:}} With the proliferation of devices at the Edge level, sharing data for centralized as well as distributed computing becomes increasingly difficult due to communication and processing bottlenecks. Decentralized computing must be able to seamlessly cater to the billions of devices that contribute data or computing resources without degrading the device performance as well as congesting the network.

    \item \textit{\textbf{Privacy and Security:}} Malicious adversarial actions against intelligent systems can encompass both the data as well as the learning framework. Deep Learning-based AI systems carry their own inherent risks \cite{Security_Akhtar_2018}, \cite{Security_Papernot_2016} , \cite{Security_Yuan_2019}. AI in a distributed setting exposes the learning architecture to a certain degree if the learning parameters are being shared. These could then be used for adversarial actions against the learners leading to degraded performance and crippling critical automated systems. Furthermore, maintaining data integrity as well as prohibiting the exploitation of the associated metadata are challenges faced whenever data is shared over the network.  This challenge in addition to the privacy risk to the data makes Edge Intelligent operation \cite{Security_XiaoJia_2019} \cite{Security_Chen_2019} more difficult.

   \item \textit{\textbf{Reliability:}} Conventional Deep Neural Networks (DNNs) do not cater to reliability issues. However, in a distributed setting, reliability guarantees need to be established for obtaining correct loss in addition to the fact that the state of the network affects the performance of the edge computing scenarios \cite{Reliability_Deng_2020}. There exists an algorithmic challenge for implementing Edge Intelligence in the presence of network (e.g. packet error and dropout, congestion) and client (e.g. offline, busy, adversarial environment) reliability issues both for training and inference process. Emergence of 5G, Ultra-Reliable-Low Latency (URLLC) networks offer promising avenues \cite{Reliability_Park2019}, however the learning frameworks themselves need to be adapted to fully utilize the benefits of these modern communication technologies.   

    \item \textit{\textbf{Inference Transparency:}} Critical learning applications for domains such as health, finance, security among others require interpretability \cite{lipton2018mythos} "as an important element of engendering trust in the inference system". Interpretability, however, is yet to have a precise definition in terms of machine learning though \cite{molnar2019}, \cite{doshi2017towards} have presented a general framework for gauging the underlying dynamics of learned decision making. \cite{Chakraborty_Tomsett} presents an elaborate survey of the existing works categorizing them under Model transparency and Model functionality for Deep Learning. A distributed setting, however, offers a unique scenario where the model transparency not only causes additional constraints on the model and operating costs, it constitutes an entirely different challenge due to the environment and nature of distributed models themselves. The first reported case of the hazards of opaque learning systems and their impact on real-life scenarios was recently reported as the wrongful arrest by Detroit Police due to a faulty match by the facial recognition system. It is imperative thus that transparency is integrated into the learning models and Edge Intelligence must adhere to the same.

    \item \textit{\textbf{Device Diversity in Edge: }}The edge levels house a myriad of clients with distinct storage and processing capabilities. The devices farther away from data origins are more capable both in terms of memory and computational power with central servers being the most resourceful devices. Furthermore, the networks employ multiple communication links and protocols which results in a highly diverse environment. These device and network characteristics exacerbate the difficulty of developing a collaborative structure for learning. Any Edge Intelligent system must therefore possess inherent tolerance for this heterogeneity. Furthermore, the designed frameworks should consider the nature (e.g. memory, processing, energy, static or mobile, data availability) of the platforms they are intended for. Platforms such as mobiles and autonomous vehicles have relatively better processing power as compared to IoT or Smart Home sensor networks. A DNN designed for IoT will not be optimal for a mobile setting and vice versa. It is therefore imperative that the learning systems consider the characteristics of their intended platforms as a tunable parameter.  
\end{enumerate}
\section{Deep Learning for Edge Intelligence}
AI has revolutionized the digital services relying on innovative new features and novel experiences such as smart home, self-driving cars, and AR/VR applications. This surge of data-driven applications has seen intelligence becoming an essential component for keeping up with growing emphasis on customized experiences while fulfilling the contrasting Service Level Requirements (SLR). Edge Intelligence is fast emerging as one of the most viable options to keep at par with this thriving digital world without congesting the core network. Within the learning paradigms, Deep Learning has rapidly evolved to deliver the best performance in some domains with major applications on the user end. \\

Deep Learning (DL) \cite{lecun2015deep} has found widespread applications in statistical learning tasks. Inspired by the anatomy of the human brain, Deep Neural Networks can learn deep representations of data by just observing large volumes of data itself. DNNs have achieved unprecedented success in computer vision for both images and videos, natural language processing \cite{otter2020survey}, medical diagnosis \cite{litjens2017survey} and security applications. Convolutional Neural Networks (CNNs), Recurrent Neural Networks (RNNs) including Long-Short Term Memory (LSTM) Networks, AutoEncoders, Generative Adversarial Networks (GANs), Deep Reinforcement Learning (DRL) \cite{liu2017survey} and their variants are some of the most commonly employed DL schemes.

\subsection{Deep Edge Intelligence (DEI)}
Traditionally, Deep Learning has been applied in a centralized environment owing to resource constraints and data requirements. Deep Learning-based Edge Intelligence or Deep Edge Intelligence(DEI) encompasses executing DL models The Edge Intelligence resulting from the confluence of Deep Learning and Edge Computing however demands a careful consideration of the operating requirements to maintain a delicate balance between often conflicting requirements already discussed in Section-\ref{EI}. We term this emerging domain as Deep Edge Intelligence (DEI). Contrary to conventional Deep Learning, environment characteristics including device capability, network capacity/mode and acceptable performance thresholds form key constraints at the outset of the design process for DEI.
\begin{center}
\begin{table*}[htb]
\caption{Summary of Related Surveys for Deep Edge Intelligence (DEI)}
\label{survey-table}
{
\small
\hfill{}
\def\arraystretch{1.5}%
\begin{tabular}{|p{0.7cm}|p{4cm}|p{6cm}|p{5cm}|}
\hline
\centering \textbf{Ref} & \centering \textbf{Scope} & \centering \textbf{Highlights} & \textbf{Major Challenges Identified}\\
\hline
\cite{Reliability_Deng_2020} & DEI based Edge-operations management and applications &
\begin{itemize} 
\item AI based network and resource management operations for EC
\item AI applications feasible for deployment in Edge environment
\end{itemize}
 & \begin{itemize}
     \item DNN model finalization
     \item Data management
 \end{itemize} \\
\hline
\cite{Zhou_Chen_Li_Zeng_Luo_Zhang_2019} &
DEI techniques for optimized training and inferences & 
\begin{itemize}
    \item DEI performance indicators
    \item DEI Techniques for inference and training
\end{itemize} &
\begin{itemize}
    \item DEI development platforms
    \item DNN and device limitations
    \item Computation aware management
\end{itemize}\\
\hline
\cite{Wang_Zhang_Wang_Ma_Liu_2020} & DEI Applications &
\begin{itemize}
    \item EI applications for autonomous structures and operations of the future
    \item Including driver-less cars, smart grids, homes and cities
\end{itemize}& 
\begin{itemize}
    \item Consolidation of emerging DEI enablers for optimal performance
\end{itemize}\\
\hline
\cite{Chen_Ran_2019} & Modes of DEI deployment across Edge nodes for training and inference & 
\begin{itemize}
    \item Identify various scenarios for optimal execution of DEI schemes on end-devices
    \item Model Splitting between devices and edge nodes and cloud servers 
\end{itemize} &
\begin{itemize}
    \item Resource management
    \item DEI benchmarking
    \item Convergence of DEI with network abstraction schemes
\end{itemize} \\
\hline 
\cite{Marchisio_Hanif_Khalid_Plastiras_Kyrkou_Theocharides_Shafique_2019}  & Optimized DEI &
\begin{itemize}
    \item Hardware, software and run-time optimization
    \item DNN security in edge environment
\end{itemize}& 
\begin{itemize}
    \item Joint Hardware-Software optimization for DNNs
    \item Hardware-aware DNN Tuning
    \item Conditional Computation
\end{itemize}\\
\hline
\cite{Wang_Han_Leung_Niyato_Yan_Chen_2020} & Explore DEI research for training and inference for both "Intelligent Edge" and applications. &
\begin{itemize}
    \item Application settings for both Intelligent Edge through automated caching, computation offloading and management operations
    \item DEI scenarios for both inference and training in addition to supporting hardware
    \item Software support for DEI and model design frameworks
\end{itemize} &
\begin{itemize}
    \item Redefining of DNN performance criteria for DEI
    \item Joint DNN optimization for training and inference
    \item widespread adoption of Intelligent Edge and emphasis on Transfer Learning.
\end{itemize}\\
\hline
\cite{Xu_Li_Li_Su_Tarkoma_Hui_2020} & DEI for caching and offloading in addition to training and inference & 
\begin{itemize}
    \item Identify core areas being utilized for each of the four domains of focus
    \item Modes and acceleration of training and inference
    \item Edge optimized model design
    \item Communication and model compression in addition to cache operation deployment
\end{itemize} &
\begin{itemize}
    \item Data disparity and uniform accessibility
    \item Pitfalls of deploying centrally trained models
    \item Privacy and security issues
\end{itemize}\\
\hline
\end{tabular}}
\hfill{}
\label{tb:tablename}
\end{table*}
\end{center}

\subsection{Deep Edge Intelligent Configurations}
As with all machine learning frameworks, Deep Learning also involves two distinctive phases:
\begin{enumerate*}[label= \emph{\alph*)}]
    \item \emph{Training} and
    \item \emph{Inference}
\end{enumerate*}
The modalities involved in both of these processes operating in a distributed environment significantly alter the way DEI problems are formulated. \emph{Training} requires data aggregation and computing resources to enable a DNN to learn. Aggregating data from a distributed environment is bound to result in considerable communication and time costs in addition to posing a significant risk to user privacy and load on the cloud network. Inference, on the other hand, requires that input be transmitted to the cloud at the cost of latency and privacy while a consistent memory footprint is required for the DNN at the cloud level. To address these issues, DEI resorts to collaborative operations spread among various elements at all levels of the network. In principle, such models can be thought of as operating in a 
\begin{enumerate*}[label = \emph{\alph*)}]
    \item \emph{Centralized (Cloud-Centric) Mode} 
    \item \emph{Decentralized (Edge Centric) Mode} 
    \item \emph{Hybrid (Joint Cloud-Edge) Mode}.
\end{enumerate*}

A summary of the surveys on the prevalent research directions for harmonizing Deep Learning for Edge operations is provided in Table-\ref{survey-table}. The areas covered under these works range from application to optimization centric and offer many prospective avenues for DEI. \cite{Reliability_Deng_2020} classifies the AI solutions in the Edge environment as AI for Edge for managing network operations and AI on edge for various AI applications operating at Edge networks. \cite{Zhou_Chen_Li_Zeng_Luo_Zhang_2019} provides an overview of the existing literature according to training and inference operations providing performance metrics. They further characterize these architectural configurations according to multiple training and inference scenarios distributed between various levels of the network. \cite{Wang_Zhang_Wang_Ma_Liu_2020} provides a review of the application-specific Deep Learning approaches for Edge Intelligence. \cite{Chen_Ran_2019} elucidates on the modes of Deep Learning operation for Edge Intelligence while \cite{Marchisio_Hanif_Khalid_Plastiras_Kyrkou_Theocharides_Shafique_2019} discusses current works on software and hardware optimization for Deep Learning in a distributed setting. \cite{Wang_Han_Leung_Niyato_Yan_Chen_2020} presents a very comprehensive overview merging various aspects of Deep Learning based solutions for improved communications, network operations and implementations for edge networks. Finally, \cite{Xu_Li_Li_Su_Tarkoma_Hui_2020} provides an extensive review extending the operational classification of the existing Edge Intelligence literature to include caching and offloading in addition to training and inference. 

\section{Deep Edge Intelligence (DEI):\\Enabling Mechanisms}\label{enabling}
\begin{figure}[ht]
\centerline{\includegraphics[scale = 0.135]{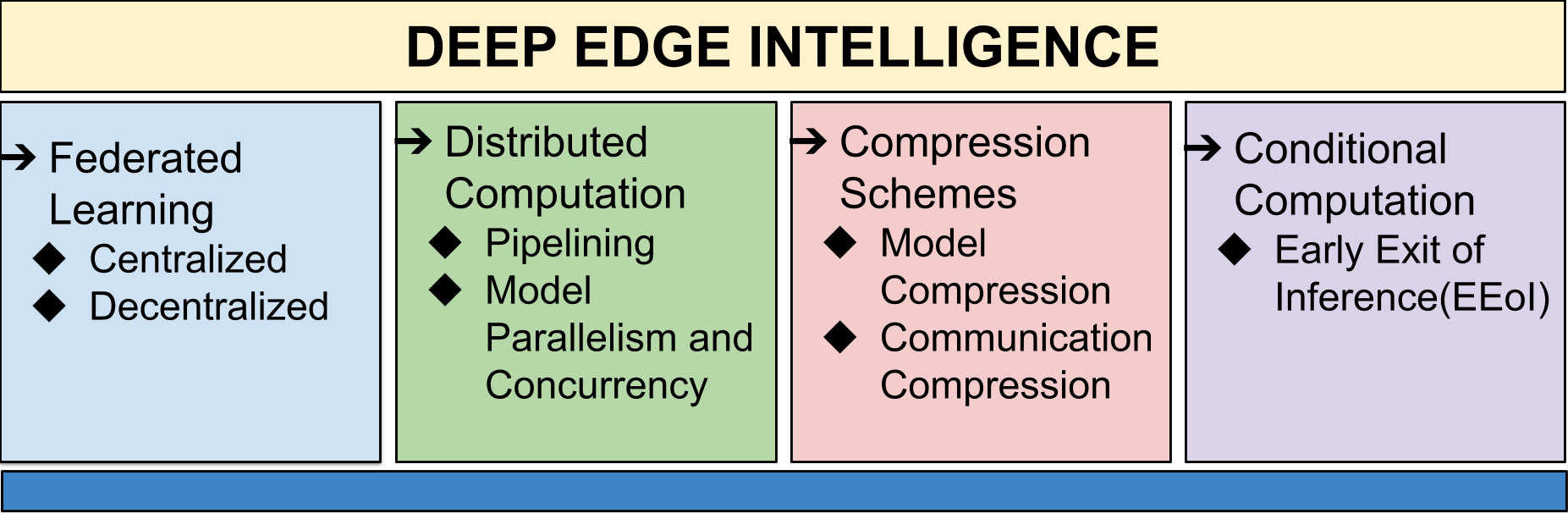}}
\caption{Deep Edge Intelligence domains with major sub-domains}
\label{fig:DEI}
\end{figure}

The realization of Deep Edge Intelligence has been made possible by some key enabling factors. These techniques are more often applied in conjunction with each other to obtain optimal performance for various settings. There also exists a considerable overlap in scenarios in which these techniques are jointly applied for allowing inter-disciplinary solutions to the challenges of DEI. These key enabling mechanisms include:
\begin{enumerate*}[label = \emph{\alph*)}]
\item \emph{Federated Learning},
\item \emph{Compression Schemes},
\item \emph{Distributed Computation} and
\item \emph{Conditional Computation}
\end{enumerate*}
as shown in Fig-\ref{fig:DEI}. Various works have elucidated on these research areas considering their usage implications in training or inference processes or their utilization in model design or modifications. Automating operations or intelligent edge applications are also ways under which these domains have been discussed. However, these factors have always contributed in one form or the other in all settings which is why we focus holistically on the key concepts which are rapidly enabling the advancement of DEI.
\subsection{Federated Learning}
\begin{figure}[htbp]
\centerline{\includegraphics[scale = 0.1]{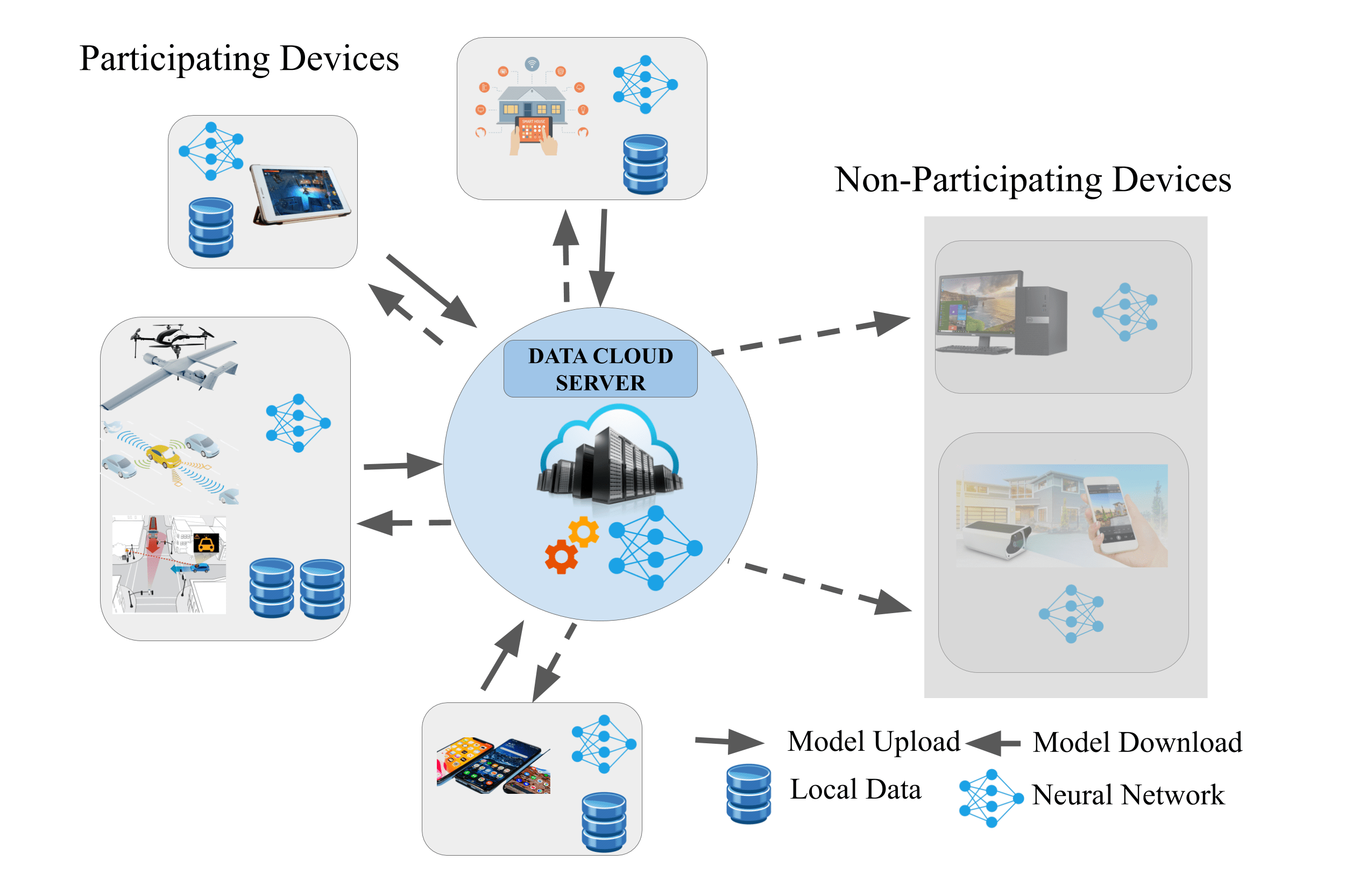}}
\caption{Centralized Federated Learning}
\label{centfl}
\end{figure}

Edge Intelligence tries to leverage the resources available mostly at the edge level of the network. However, this domain has a heterogeneous nature both in terms of the device and the communication protocols that link these devices. Federated learning, proposed in \cite{McMahan_Moore_Ramage_Hampson_Arcas_2017}, enables these entities (clients) to learn collaboratively while being coordinated by a central server without ever exchanging raw data. This mechanism decentralizes the learning process by enabling clients to start from a common DNN and train on the locally available data. Federated Learning ensures the privacy of the users since no data is communicated between the clients during the training process. Subsequently, multiple clients are then solicited by the central server to share their parameters which are then aggregated by the server itself (Fig-\ref{centfl}). The updated model is then broadcast to the clients which then repeat the process until convergence. The inference process takes place on the client itself and thus data is never transmitted over the network which ensures user and client privacy. \cite{FL_Kairouz} in a comprehensive survey classifies the challenges for federated learning dividing the issues as either "Algorithmic" or "Practical" in nature. Communication and model optimization are the two key parameters that are used to gauge the effectiveness of an FL algorithm. The other key issues which dictate the current research avenues in federated setting are:\\
\paragraph{Non-IID and Imbalanced Data} Various works have tried to build on the Spatio-temporal relationships present in the data, particularly for a geographical vicinity. Most such works deal with Edge Intelligent Caching, however, this cannot be used as a generally acceptable property of data. As no prior information is available about the data being held at various clients, the data is thus characterized as non-Independent and Identically Distributed (non-IID) as well assumed to be class imbalanced. Furthermore, differential privacy may be ensured by restricting participation by overly eager clients or adding noise to the data at the clients, thus preventing memorization.
\paragraph{Inconsistent participation}  Adverse network conditions, as well as offline devices, are catered by the central server by orchestrating a pool of devices to share their parameters. Additionally, the access to training data as well as processing capability across the edge is non-uniform. The aggregation protocols should be robust to client/network outages or degraded contributions in a federated setting
\paragraph{Privacy and Security} Federated learning maintains privacy by communicating only model parameters or updates. Furthermore, application of secure aggregation aims to add another layer of security on the communicated data. Additionally, However, malicious action against the participating devices in form of poisoning data, update corruptions can compromise the aggregation process ultimately affecting all the clients. Thus, there remains a need to introduce trust guarantees for the clients to ensure secure FL.
\paragraph{Synchronous vs Asynchronous Operation} A typical federated setting entails some level of synchronization to be placed to ensure timely updates for the client model. However, varying network conditions, as well as client capabilities and presence are conditions that can seriously degrade synchronous federated learning. While asynchronous orchestration of clients in a federated setting proves robust to the challenges caused by adverse network and communication conditions, such designs also cause higher latency and may give rise to convergence issues. As an example, \cite{CC_koloskova2019decentralized} establish that the consensus mechanisms implemented through standard gossip algorithms in the presence of compressed communication do not converge.\\
\newline
\cite{FL_McMahan_Moore_Ramage_Hampson_Arcas_2017} while introducing Federated Learning identifies data related challenges as key issues for a federated setting. Experimental results indicate that the proposed \emph{Federated-Averaging} possesses a degree of robustness to the class-imbalanced and non-IID data. Then, \cite{FL_Stich_2019} provides theoretical guarantees for \emph{local Stochastic Gradient Descent (SGD)} running on convex objective functions on batch learning problems, operating on individual clients and averaged on a central server. \cite{FL_Chen_Sun_Jin_2019} presents Temporarily Weighted Aggregation Federated Learning (TWAFL): an asynchronous mechanism for aggregating and updating parameters of shallow and deep layers of DNNs at different frequencies while aggregating the model parameters using temporal weighting. The authors report speeding up of the convergence process assigning higher weights to recent updates. \cite{FL_Wang_Han_Wang_Zhao_Chen_Chen_2019} proposes In-Edge AI using Deep Reinforcement Learning in a federated setting for caching and computation offloading. The paper uses a Double Deep Q-Network for both caching and offloading operations being used in federated settings and reports performance on par with centralized DRL algorithms. An attempt to mitigate the reliability issues associated with central aggregation has been conducted in \cite{FL_Kim_Park_Bennis_Kim_2020} where BlockFL, a blockchained Federated Learning architecture, is proposed. The model updates computed locally by the clients are uploaded to the associated miner which after performing the Proof of Work (PoW) uploads the local updates to the distributed ledger. The global update is computed at the individual clients accessing the distributed ledger. This decentralization, however, is achieved at the cost of greater latency during the training process.

\subsection{Compression Schemes} Compression schemes are aimed at reducing the communication costs and memory footprint costs in DEI settings. To this end, the major applications of compression lie in  two areas: 
\begin{enumerate*}[label = \emph{\alph*)}]
    \item \emph{Model Compression} and
    \item \emph{Communication Compression}\\
\end{enumerate*}
\paragraph{Model Compression} The parameters of DNN models can vary from thousands to hundreds of millions of parameters. Model compression aims to compress to reduce the storage requirements of the DNN models as well as to speed up the training process without degrading the accuracy of predictions. Model compression can be achieved by five major methods \cite{Xu_Li_Li_Su_Tarkoma_Hui_2020}: Dimensionality reduction, pruning and sparsification, quantization, knowledge distillation and layer design. Dimensionality reduction is achieved through low-rank approximation and matrix factorization which aim to construct reduced rank representative matrices or decompose the dense original matrices such as weights of DNN layers. However, this process requires fine-tuning of the extracted matrices often accompanied by a loss of accuracy. Network pruning entails the elimination of insignificant parameters within a DNN leading to a sparse network. Pruning can be applied to both individual weights (fine-grained) or entire layers (coarse-grained). Once sparsification is achieved, the network is retrained while keeping the pruned elements frozen. The requirement of deploying high precision DNNs is evaluated and downgraded accordingly during the network quantization process. The existing redundancies in the DNNs are exploited for both eliminating and reducing precision while obtaining acceptable performance. Knowledge distillation, inspired from transfer learning, entails the construction of a smaller DNN which mimics the behavior of a larger DNN. The process involves creating a deeper teacher model which is then used to train the more compact student model. Layer design is aimed at introducing compactness in the DNN at the architectural level. Pooling has been one of the ways conventional deep networks have tried to introduce compactness in the network architecture. The principles of model compression apply to multiple DNN settings and have found active use cases in federated and distributed settings. \\

\cite{MC_maji2017adapt} explores the application of a low-rank approximated Convolution Neural Network (CNN) model in inference generation while deployed in an IoT network. Convolutional layers have benefited from compact design involving smaller filter size, reduced channel depth and downsampling, all aspects which were combined in Squeezenet to achieve a 50x reduction in the number of parameters with a Top-5 accuracy of 83\%, which is comparable to that of AlexNet. \cite{MC_shafiee2017squishednets} further builds on this and explore the feasibility of deployment of their proposed model on edge devices. p\textbf{R}ivate m\textbf{O}del compressio\textbf{N} fr\textbf{A}mework, RONA, a knowledge distillation framework is proposed in \cite{MC_wang2019private}. The presented framework emphasizes on differential privacy and explore its performance on a mobile platform. \cite{MC_Yang_Chen_Sze_2017} presents an energy-aware pruning mechanism for CNNs where the energy consumption of the respective layers are evaluated. The pruning is conducted on the largest layers with the highest energy requirements and the network is retrained after each pruning round. Finally, the entire model is globally fine-tuned to achieve the best energy-accuracy trade-off.\\

\paragraph{Communication Compression} At the other end of the spectrum are the applications of compression techniques on the communication involved in operating DEI. Parameter sharing during the training process in federated deep learning and feature uploads for both inference and distributed are some of the key beneficiaries of communication compression. Dimensionality reduction of the DNNs results in structured updates being learned to consist of a lesser number of parameters. On the other hand, the updates being shared by the clients are themselves compressed using quantization and various coding schemes. In addition to the communication itself, the frequency of communication also becomes critical at higher scales involving a large number of devices. Gossip training provides an alternative to centralized training, instead relying on gossip algorithms to achieve faster convergence, reduced upstream communication cost and complete decentralization. Furthermore, adaptive client selection for participation in the aggregation is another active area of research that can yield reduced communication costs. Following such a mechanism, the clients with valuable contributions are selected for aggregation in a round. \\

Deep Gradient Compression (DGC) by \cite{lin2017deep} has reported that 99.9\% of the gradients exchanged in distributed SGD is redundant to successfully employ gradient compression for distributed training. \cite{daily2018gossipgrad} shows that without appropriate customization, regular gossip training algorithms result in poor convergence and far greater communication costs. The authors instead propose GossipGrad for better scalability reducing the overall communication complexity from $\Theta(log(p)$ to $O(1)$. To further reduce edge-cloud communication in a federated setting,  \cite{CC_koloskova2019decentralized} presents CHOCO-SGD as a communication efficient distributed Stochastic Gradient Descent (SGD) and aggregation method. They also present CHOCO-Gossip as a converging gossip algorithm for the distributed average consensus problem under compressed communication conditions. 

\subsection{Distributed Computation}
Varying combinations of limited power, memory and computation resources act as major bottlenecks for edge devices while sending inputs upstream can cause network congestion, adversely affecting privacy and latency. Distributed computation sees a collaboration of available processing resources in achieving adequate processing power. Distributed computations can be categorized into two groups:
\begin{enumerate*}[label = \emph{\alph*)}]
\item \emph{Pipelining} and
\item \emph{Model Parallelism and Concurrency.}
\end{enumerate*}
\\
\paragraph{Pipelining} Pipelining \cite{ben2019demystifying} serves as an effective tool in distributing specific DNN computations or layers among various on-device processors or other nodes across the network. The former method is used in a general multi-processor environment to speed up processing while the latter is referred to as \emph{(Model Segmentation)} in the context of DEI. Implementing the segmented structure of a DNN can result in reduced energy costs and improved inference latency. On the other hand, \emph{Computation Offloading} can be considered as an offshoot of model segmentation. However, computation offloading is more adaptive in terms of collaborating with nodes based on the prevalent network conditions and various requirements. Model Segmentation and Computation Offloading are frameworks that reflect a collaborative processing among various devices spread across the network hierarchy. 
\begin{itemize}
\item \emph{Model Segmentation}:  Model Segmentation is implemented by partitioning the DNN layers among various devices across the levels of the network as depicted in Fig-\ref{fig:segment} . This enables the architecture to leverage collaborative operation using the resource-constrained devices at the edge. Various blocks or layers are housed at various levels across the network. With split models, only the learned features instead of the inputs are shared. Additionally, features communicated by the preceding portion of the DNN to the subsequent layers are much smaller than the input size and thus has the potential to reduce inference latency. Segmented models also achieve improved latency and energy efficiency. 
\begin{center}
\begin{figure}[htbp]
    \centering
    \includegraphics[scale = 0.1]{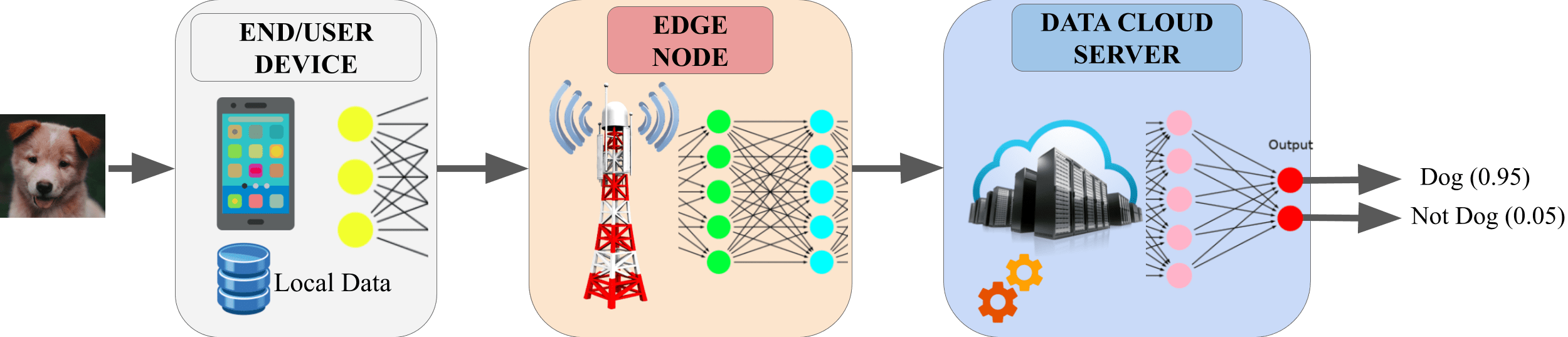}
    \caption{Model Segmentation}
    \label{fig:segment}
\end{figure}
\end{center}
\item \emph{Computation Offloading}: Computation Offloading is an optimization-based method which enables the tasks to be computed off-device. Offloading techniques enable the complete or partial offloading of DNN computations deeper along the network. However, the decision to offload computation needs to consider trade-offs between various factors including but not limited to inference latency and accuracy, network status as well as target device resources and scheduling. Computation offloading can be achieved in the following configurations \cite{Wang_Han_Leung_Niyato_Yan_Chen_2020}:\\
\textbf{\emph{Partial Offloading}} takes place when some of the tasks are uploaded to other nodes and cloud. \textbf{\emph{Horizontal Offloading}} allocates concurrent tasks to edge devices which then communicate their results to the cloud server. A natural extension of such an offloading scheme for the highly resource-limited end-devices is to only perform input pre-processing to produce a Region of Interest (RoI) which is then fed to the classifier/s. \textbf{\emph{Vertical Segmentation}} entails that the tasks are allocated hierarchically along the network upstream. Each preceding result becomes an input to the subsequent stage.\\
\end{itemize}
\paragraph{Model Parallelism and Concurrency}: The systems present across the network have a highly heterogeneous nature. The smart mobile devices have particularly seen a proliferation of multi-core processor systems which may be utilized for an edge intelligent environment. Model parallelism divides the computation among various on-device processors. For a convolutional layer in a DNN, filters in the depth dimensions can be assigned to separate processors to run simultaneously. Attempts have also been made to configure DNNs to maximize parallelism and minimize communication by introducing redundant operations within the DNN architecture \cite{muller1994neural}. Model concurrency, on the other hand, overlaps operations such as forward and backward pass and weight updates by executing them simultaneously.\\

GPipe \cite{Huang_Cheng_Bapna_2019} pertains to model segmentation where a DNN partitioned into \emph{K} cells. Each of these cells can then be placed on a different accelerator which is then trained on micro-batches. \cite{kang2017neurosurgeon} proposes \emph{Neurosurgeon}, a framework to partition layers across the network based on their compute and data characteristics. \cite{hadidi2019collaborative} explores the feasibility of model parallelism techniques for highly resource-constrained IoT networks and propose an evenly distributed data processing pipeline. JointDNN proposed in \cite{eshratifar2019jointdnn} presents a collaborative learning method between edge and cloud while also employing compressed communication. Finally, DNNs have also been converted into Directed Acyclic Graphs (DAG), which are then simplified using pruning and the process is repeated to eventually yield a much simpler graph network. In \cite{hu2019dynamic}, the authors present the Dynamic Adaptive Surgery Scheme (DADS) to segment DNNs under various network conditions by treating inference minimization as a min-cut problem.  

\subsection{Conditional Computation}
DNNs with greater depth have displayed the ability to learn deeper representations of the input. However, this depth is often accompanied by prohibitive computation, latency and memory costs. Conditional computation mechanisms try to establish the best trade-off between accuracy and a combination of constraints discussed in Section-\ref{constraints} which they achieve by exploiting redundancies. A conditionally computing learning system may resort to selectively deactivating certain network operations if the execution does not yield considerable improvement in the inference quality. This mechanism is called \emph{"Early Exit of Inference {EEoI}"}. In terms of a DNN, it can be compared to the drop out layers; the only difference being the case of EEoI, entire computation blocks are restricted from operating. The principle of conditional computation may also be extended to select the most valuable contributors in a Federated Learning setting. In such a case, only those clients may become part of the aggregation which have a significant impact on the network learning or performance. Another application of EEoI lies in the domain of \emph{Edge Caching} which involves skipping DNN execution if the features of the current input resemble those of a cached inference. \\
    \paragraph*{\textbf{Early Exit of Inference}} Early Exit of Inference (EEoI) has been established that the shallow layers of a DNN learn more general features whereas deeper layers learn more specific ones. Early Exit of Inference, based on this observation, is another mechanism to obtain a balance between inference latency and accuracy. In such cases, Early Exit of Inference (EEoI) enhances efficiency by allowing the inferences to be generated by earlier layers instead of having the input to traverse the entire depth of the DNN. The suggested mechanism is employed if the inference can be generated with high confidence. EEoI has also been combined with Model Segmentation to allow computation offloading in cases where the inference does not meet the required thresholds. A major benefit of EEoI when compared with Model Segmentation and Offloading Computation is its potential for reduced communication overhead and improved inference latency.\\
    EEoI has also found its application in the context of \emph{Edge Caching} which is based on identifying "redundancy of requests" \cite{Xu_Li_Li_Su_Tarkoma_Hui_2020}. Edge Caching is inspired by the same mechanism used to speed up memory access in computing systems. The efficiency of Edge Caching systems is conditional upon effectively identifying the redundancies in the communication or computation.  Edge Caching has conventionally been categorized into \emph{"What", "Where" and "How" to cache} at the Edge Network. Attempts to answer these through the application of various DL applications have been done among others in \cite{cache_chang2018learn}, \cite{cache_Chen_He_Liu_Lan_Chung_Mao_2019}, \cite{cache_rathore2019deepcachnet}. These and other related works have managed to explore these questions by centralized learning predicting popular content based on the features of the user requests or training a DNN using the outputs from heuristic algorithms. These methods certainly improve the caching performance and yield high Quality of Experience (QoE). However, the facilitation of DEI deployment for Edge Caching is achieved by addressing the problem of "How to Access" the cache. DNNs deployed in a geographically co-located manner are likely to receive similar requests. Caching the results of previous inferences can enable future inputs to skip DNN execution if their features match the features of the stored inference with a certain confidence. In this manner, EEoI is achieved in case of a successful cache 'hit.    
    
    \cite{zeng2019boomerang} proposes 'Boomerang', a cooperative DNN framework for IoT employing EEoI mechanism by training with multiple exit points. Shallow-Deep Networks by \cite{kaya2019shallow} introduces the concept of 'Overthinking' in DNNs to elaborate redundant and wasteful computations in DNNs particularly CNNs. The authors term their use of EEoI as Internal Classifiers (ICs) which they introduce along various stages of the DNN to combined shallow and deeper layers as required. \cite{drolia2017cachier} proposes \emph{Cachier}, which stores the inferences as well as the corresponding inputs features at the Edge Nodes. An offline machine learning model trained on features from possible queries is then used to find the best match for the features from the current input. CacheNet, proposed in \cite{fang2020cachenet}, is a mechanism to cache DL models on end devices as well as edge nodes. The end device houses multiple smaller submodels trained on knowledge partitions whereas baseline models are kept at the edge device. The end-device uses a 'hint' mechanism to select the model most suitable for the input data which in case of a 'miss' uses the edge model for inference.
\section{Open Challenges and \\Future Research Directions}\label{open}
\begin{figure}
    \centering
    \includegraphics[scale = 0.35]{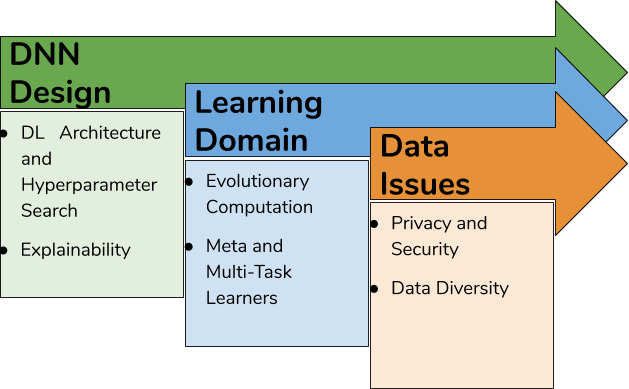}
    \caption{Some open research directions for DEI}
    \label{fig:challenge}
\end{figure}

DEI offers a powerful solution for ensuring the ubiquity of AI for the current networks with a considerable space for future-proofing. Despite the obvious benefits, there still exist several open challenges and research directions that may be explored to enhance the effectiveness of DEI. The open challenges can be categorized under three broad categories as depicted in Fig- \ref{fig:challenge}:\\
\begin{enumerate}
    \item DNN Design
    \item Learning Domain
    \item Data Issues\\
\end{enumerate}
Each of these domains presents its own unique set of challenges addressing which are essential for DEI to flourish. We present some research avenues and open challenges under each of these areas in the subsequent pages. 

\subsection{DNN Design}
The DNN architectures for DEI should be designed with attention to the operating edge environment to perform optimally. Special emphasis should be laid on adapting the models to the edge constraints which in turn brings its own set of challenges. Going forward, \emph{DL Architecture and Hyperparameter Search}  and  \emph{Explainability} are two significant challenges for DEI.\\

\paragraph{DL Architecture and Hyperparameter Search}
DL is being employed to address increasingly challenging problems. Since the early days of MNIST classification problems, the models have evolved to become deeper and more sophisticated in terms of the layers and the connections they employ. As elaborated in Section-\ref{EI}, the edge network depicts a significant diversity in devices as well as data (IID vs. non-IID). DNN architecture designs need to cater for these additional conditions for an optimal performance. The DNN architecture engineering for such a varied environment may be excessively challenging and time consuming for traditional human efforts. The success of Automated Machine Learning (AutoML) which allows for automated DNN architecture search is evident by the fact that the current ILSVRC winner \cite{touvron2020fixing} is based on an architecture refined by AutoML. Although AutoML frameworks such as those offered by Google and Microsoft Azure offer powerful tools, they explore model design space for centralized learning settings. Model segmentation, concurrency and federation are some aspects essential to DEI which are currently absent from such automated search. \cite{he2018amc} and \cite{tan2019mnasnet} propose an AutoML framework for refining and designing mobile-based DNN architectures respectively. However, the approach is still limited to a family of devices and AutoML may be required to cater for the additional constraints offered by the edge environment. AutoML with the incorporation of edge constraints may be employed to accelerate designing and testing DNNs for DEI.\\

\paragraph{Explainability} 
The concept of model explainability has been briefly stated in Section-\ref{EI}. However, the prospect of interpretability in the context of DEI is itself a very challenging topic. Inference explainability for DL has garnered considerable importance as a means to provide context to the statistical metrics used to gauge the efficacy of the learner network. explainability for conventional models has so far been achieved by using consolidating information across the depth of the network. In conventional DNNs, the notion explainability has been addressed through "Attention Mechanisms" \cite{Vaswani_Shazeer_Parmar_Uszkoreit_Jones_Gomez_Kaiser_Polosukhin_2017} \cite{Jetley_Lord_Lee_Torr_2018} or inference-time/post-hoc processes such Gradient Weighted Class Activated Mapping (Grad-CAM) \cite{Selvaraju_Cogswell_Das_Vedantam_Parikh_Batra_2019}. For DEI, however, the shallow nature and segmentation are key challenges that need to be catered for to incorporate interpretability. Furthermore, coming up with mechanisms that provide explainability without considerable communication and computation costs is a highly challenging prospect.

\subsection{Learning Domain}
Learning algorithms dictate the effectiveness of DEI. However, on-device resources at the edge network are scarce and the homogeneity, availability and quality of data all are highly conditional. A considerable amount of research is being carried out to address the former issue in the form of distributed computation schemes and specialized hardware like NVIDIA Tensor Cores, Tensor Processing Units and Application Specific Integrated Circuits (ASICs) \cite{ben2019demystifying},however, limited research has been carried out in exploring algorithmic alternatives. \emph{Evolutionary Computation} and \emph{Meta and Multi-Task Learning} are two powerful algorithmic families which have discussed below in the context of DEI as powerful alternatives for research.\\

\paragraph{Evolutionary Computation for DEI}
Actions in a network are propagated widely in the Spatio-temporal domain. With intelligent clients becoming prevalent in the emerging networks, predicting the interactions between various entities is likely to become highly complex. Google DeepMind's AlphaStar \cite{Vinyals_Babuschkin} offers an interesting case study in this perspective for an environment which is characterized by \emph{Real-time interaction, Partially observable environment, Complex action and state space} and \emph{Non-unique strategic perspectives}, all aspects which are extant to varying levels in the current and future networks. While the highlight of the original work refers to the efficacy of fusing various aspects of AI research, \cite{Arulkumaran_Cully_Togelius_2019} highlights the effective use of Population-Based Training (PBT) in the former, a central notion of the Evolutionary Computation. Additionally, PBT is both asynchronous and distributed \cite{820127} and exploits scalability to overcome prohibitive memory and time costs associated with training large single networks. Federated Learning may be considered a variant of PBT, however, the aggregation schemes merge the models without attaching considerable significance to individual conditions and performance. PBT may find significant usage in DEI environments where co-located clients may be enabled to learn competitively or collaboratively in a Game Theoretic setting especially for autonomous operations. To the best of our knowledge, very limited work has been carried out to explore the opportunity afforded for DEI by Evolutionary Computing.\\

\paragraph{Meta \& Multi-Task Learners}
The DL models in the DEI environment are much more likely to come across tasks or data they have not been trained on. Having a distinct model for each of these expected scenarios is a prohibitive prospect. Therefore, AI domains that enable generalized learning may be able to address this challenge. Meta-Learning is one such paradigm in which a learner trained on certain data can learn a new task with relatively few examples. A Meta-Learning based model-agnostic learning framework for gradient descent based learners including DL was proposed in \cite{finn2017model}. For a distributed setting a Federated-meta-learning system was proposed in \cite{Lin_Yang_Zhang_2020}. The system uses a set of nodes that collaborate in a federated fashion to learn their tasks. Once an external node is required to perform a certain task, the coordinating platform shares the model with that target node to enable it to adapt the model using a few iterations. \cite{Fallah_Mokhtari_Ozdaglar_2020} presents a meta-learning based Personalized Federated Learning. The authors propose the determination of an initialization point for the model for all the clients which enables them to learn their respective tasks in a few iterations. Similarly, Multi-Task Learning (MTL) is another domain which may prove advantageous for DEI models. The goal of MTL is to quickly learn models for multiple tasks simultaneously.  Then \cite{Smith_Chiang_Sanjabi_Talwalkar_2017} proposes Federated Multi-Task Learning in which the model weights of the clients learning their tasks, are constrained to be related.

\subsection{Data-associated Challenges}
The proliferation of AI has been possible because of the accessibility of data. With its distributed nature, DEI offers a multitude of challenges in securing and efficiently utilizing data. The risk to \emph{Privacy and Security} and the issues arising from \emph{Data Diversity} are two major issues at the heart of the challenges of the effective utilization of data. \\

\paragraph{Privacy and Security}
The emergence of data while offering numerous advantages, also holds significant risks to individual privacy and security in the hands of malicious actors. DEI requires varying levels of participation from the networked elements from minimized data sharing in Federated Learning to offloading computation all the way to the cloud. The threat to DEI is two-fold: Comprising the user privacy or the learner integrity. Ensuring privacy is currently achieved via \cite{FL_Kairouz} Secure Multi-Party Computations (MPC), Differential Privacy and Transparency. Security on the other hand may be targeted by overwhelming the device/s, corrupting data and model updates and malicious exploitation of the knowledge of the DNN models for inference evasion. A DL based unsupervised method for adversarial attack detection in the Edge Computing scenario has been proposed in \cite{Chen_Zhang_Maharjan_Alam_Wu_2019}. The suggested scheme actively learns the features of attacks in an unsupervised manner. Similarly, \cite{Bagdasaryan_Veit_Hua_Estrin_Shmatikov_2019} highlights the vulnerabilities of a Federated Learning setting. The authors envisage that a model replacement scheme where a malicious actor aims to replace the global model with a backdoored model while ensuring that the latter survives the averaging process. \cite{Mallick_Dwivedi_Kailkhura_Joshi_Han} is a representative work that highlights the problems with inference evasion in a scenario where a cat image is falsely classified as Covid-19 positive chest X-ray image by both DNNs and Bayesian Neural Network (BNN). This work refers to the problems which arise when common DL models are faced with Out-of-Distribution data. It also highlights that armed with knowledge about the DNN, malicious agents can effectively bypass DL based filtering and security measures.\\ 

\paragraph{Data Diversity}
Data plays a pivotal role in the successful execution of any learning system including DL. Traditionally, data aggregation for centralized learning resulted in maintaining a level of homogeneity. The challenges related to data for DEI on the other hand are:
\begin{enumerate}
    \item \emph{Accessibility, Quality and Bias:} With DEI becoming pervasive at the edge, a key requirement is for the DEI models to get access to the user data which would require a formal mechanism to be in place. An inherent property of this mechanism has to be ensuring the privacy of the users/data origin. Additionally, there is no existing way to gauge the quality of data being provided by various Edge entities. The data may be noisy, feature-poor, mislabelled or downright malicious. Furthermore, any data which may be used for DEI learning will be biased which must be catered for by the DEI models.
    \item \emph{Heterogeneity} The myriad of devices residing at the edge levels follow a wide array of protocols for data collection. The data accessible for DEI is very likely to be non-uniform. Attempts to bring homogeneity in this data will lose valuable information in the process which allows the DL models to learn the deep distinctive features. Another aspect of data heterogeneity is associated with the environment as well as the source where the data is being collected \cite{Xu_Li_Li_Su_Tarkoma_Hui_2020}. Environments or sensors can be overly noisy which adversely affects the data being collected. Furthermore, various sensitivity and quantization levels for sensors measuring the same quantity even in the same environment may result in large variations in the data generated. Additionally, data augmentation is commonly used for DL models to improve the learning process. In a distributed environment, augmentation across all data instances may also lead to exacerbation in the adverse properties of the data, the effects of which are likely to propagate across devices due to model sharing.
\end{enumerate}
\section{Conclusion}\label{conclusion}
This paper aims to consolidate the current trends in research on Edge Intelligence vis-a-vis Deep Learning. Due to the relative nascence of the Edge Intelligence field, we feel that there exists a considerable diversity in the nomenclature and areas being identified. During the formulation of this report, we have aimed to conceptually consolidate the existing areas currently being researched in Deep Edge Intelligence.  Initiating with an emphasis on the fact that data and devices are on track to become prohibitive for centralized computations, we introduce the concepts behind Edge Intelligence subsequently focusing on Deep Edge Intelligence (DEI). We summarize some of the key factors which govern the current and prospective areas of research among which to the best of our knowledge, 'Model Transparency' and 'Device Limitations' have not been explicitly covered previously. These enablers or their combinations have been employed in various DNN models to jointly to achieve optimal performance in a DEI setting. The major challenges for DEI vary from DNN design to the learning frameworks for efficiently dealing with the constraints of DEI. \\

It is worth noting that despite these early successes, the network and data environment is predicted to increase at breakneck pace with 6G communication envisioned to disrupt the way services are provided currently. One of the key challenges to developing lasting frameworks for DEI is to establish theoretical guarantees for DEI. Such works will be able to identify potential solutions that can persevere in the 6G era and beyond. Furthermore, better communication links will enable an exponential proliferation of heterogeneous data at the edge network. Effective learning strategies such as Multi-Task Learning, Self-Supervised learning and Meta-Learning are some of the key domains whose application in DEI can lead to more generalized learning schemes.

\Urlmuskip=0mu plus 1mu\relax
\bibliography{Deep-Edge.bib}
\bibliographystyle{IEEEtran}
\end{document}